\pdfoutput=1
\documentclass[10pt,twocolumn,letterpaper]{article}

\usepackage{cvpr}
\usepackage{times}
\usepackage{epsfig}
\usepackage{graphicx}
\usepackage{amsmath}
\usepackage{amssymb}
\usepackage{color}

\usepackage[ruled,linesnumbered,noline,commentsnumbered]{algorithm2e}
\usepackage{graphicx}  
\usepackage{subcaption}
\SetFuncSty{emph}

\DeclareMathOperator*{\argmax}{arg\,max}

\usepackage{enumitem}
\usepackage{multirow}

\definecolor{citecolor}{RGB}{119,185,0} 
\usepackage[pagebackref=true,breaklinks=true,letterpaper=true,colorlinks,citecolor=citecolor,bookmarks=false]{hyperref}

\cvprfinalcopy 

\def\cvprPaperID{1670} 

\makeatletter
\renewcommand\paragraph{\@startsection{paragraph}{4}{\z@}%
                        {1ex \@plus1ex \@minus.2ex}%
                        {-1em}%
                        {\normalfont\normalsize\bfseries}}
\makeatother
                        
\ifcvprfinal\fi
\begin{document}

\title{STEP: Spatio-Temporal Progressive Learning for Video Action Detection}

\author{
  Xitong Yang${^{1}}$\thanks{Work done during an internship at NVIDIA Research.} \quad Xiaodong Yang${^2}$ \quad Ming-Yu Liu${^2}$\\
  Fanyi Xiao${^3}$\footnotemark[1] \quad Larry Davis${^1}$ \quad Jan Kautz${^2}$\\ 
  $^1$University of Maryland, College Park ~~~$^2$NVIDIA ~~~$^3$University of California, Davis\\
}

\maketitle
\thispagestyle{empty}

\begin{abstract}
In this paper, we propose \textbf{S}patio-\textbf{TE}mporal \textbf{P}rogressive (STEP) action detector---a progressive learning framework for spatio-temporal action detection in videos.
Starting from a handful of coarse-scale proposal cuboids, our approach progressively refines the proposals towards actions over a few steps.
In this way, high-quality proposals (i.e., adhere to action movements) can be gradually obtained at later steps by leveraging the regression outputs from previous steps.
At each step, we adaptively extend the proposals in time to incorporate more related temporal context.
Compared to the prior work that performs action detection in one run, our progressive learning framework is able to naturally handle the spatial displacement within action tubes and therefore provides a more effective way for spatio-temporal modeling. 
We extensively evaluate our approach on UCF101 and AVA, and demonstrate superior detection results. 
Remarkably, we achieve mAP of 75.0\% and 18.6\% 
on the two datasets with 3 progressive steps and using respectively only 11 and 34 initial proposals.
\end{abstract}

\section{Introduction}

Spatio-temporal action detection aims to recognize the actions of interest that present in a video and localize them in both space and time.
Inspired by the advances in the field of object detection in images \cite{girshick2015fast,liu2016ssd}, most recent work approaches this task based on the standard two-stage framework:
in the first stage action proposals are produced by a region proposal algorithm or densely sampled anchors, and in the second stage the proposals are used for action classification and localization refinement.

Compared to object detection in images, spatio-temporal action detection in videos is however a more challenging problem.
New challenges arise from both of the above two stages when the temporal characteristic of videos is taken into account.
First, an action tube (i.e., a sequence of bounding boxes of action) usually involves spatial displacement over time, which introduces extra complexity for proposal generation and refinement.
Second, effective temporal modeling becomes imperative for accurate action classification, as a number of actions are only identifiable when temporal context information is available.

\begin{figure}
\begin{center}
\includegraphics[width=\columnwidth]{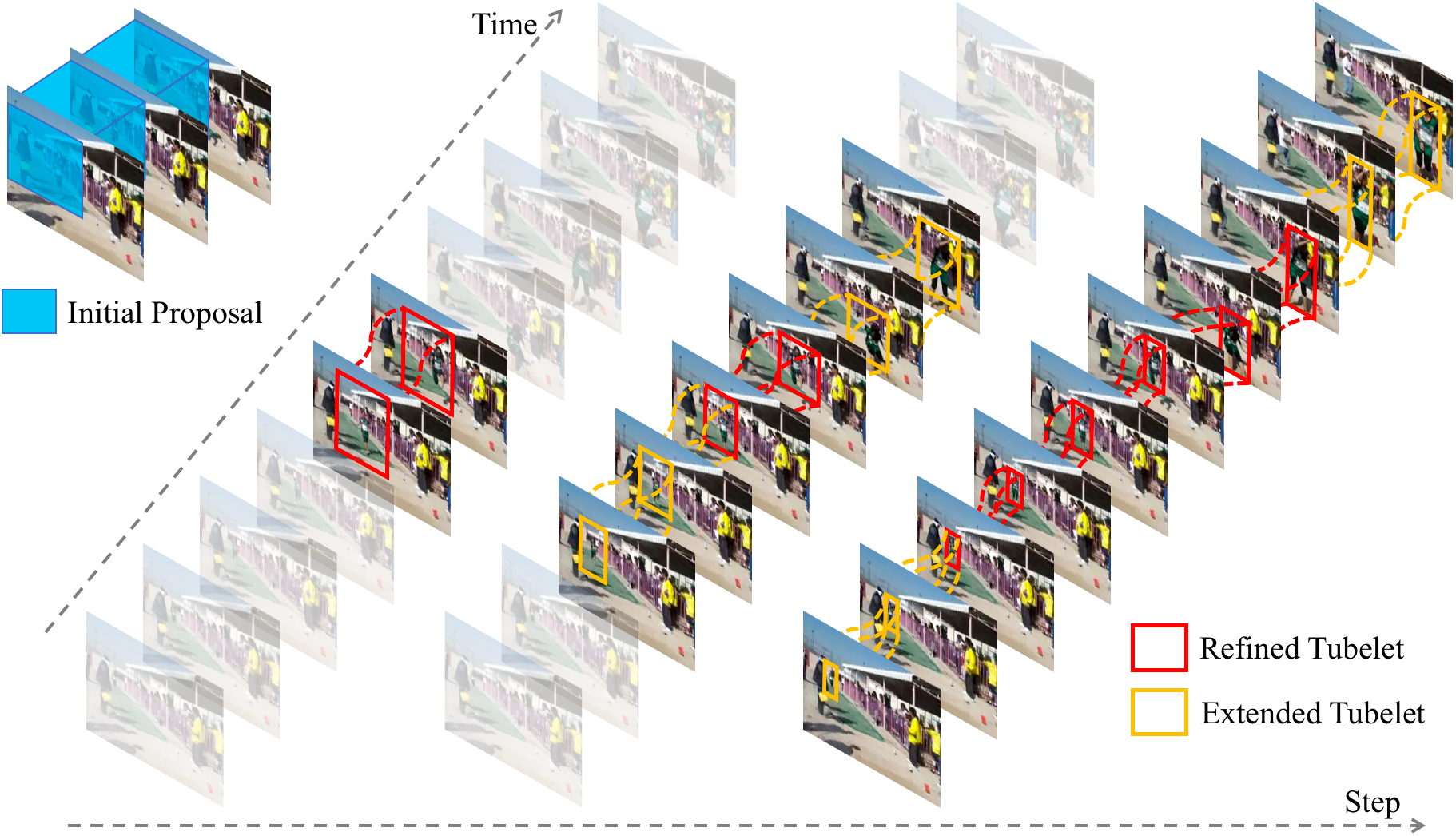}
\end{center}
\vspace{-.1in}
\caption{A schematic overview of spatio-temporal progressive learning for action detection. Starting with a coarse-scale proposal cuboid, it progressively refines the proposal towards the action, and adaptively extends the proposal to incorporate more related temporal context at each step.}  
\label{fig:1}
\end{figure}

Previous work usually exploits temporal information by performing action detection at the clip (i.e., a short video snippet) level.
For instance, \cite{gu2017ava,kalogeiton2017action} take as input a sequence of frames and output the action categories and regressed tubelets of each clip.
In order to generate action proposals, they extend 2D region proposals to 3D by replicating them over time, assuming that the spatial extent is fixed within a clip.
However, this assumption would be violated for the action tubes with large spatial displacement, in particular when the clip is long or involves rapid movement of actors or camera. 
Thus, using long cuboids directly as action proposals is not optimal, since they introduce extra noise for action classification and make action localization more challenging, if not hopeless.
Recently, there are some attempts to use adaptive proposals for action detection \cite{hou2017tube,li2018recurrent}.
However, these methods require an offline linking process to generate the proposals.

In this paper, we present a novel learning framework, \textbf{S}patio-\textbf{TE}mporal \textbf{P}rogressive (STEP) action detector, for video action detection. As illustrated in Figure~\ref{fig:1},
unlike existing methods that directly perform action detection in one run, our framework involves a multi-step optimization process that progressively refines the initial proposals towards the final solution.
Specifically, STEP consists of two components: \textbf{spatial refinement} and \textbf{temporal extension}.
Spatial refinement starts with a small number of coarse-scale proposals and updates them iteratively to better classify and localize action regions.
We carry out the multiple steps in a sequential order, where the outputs of one step are used as the proposals for next step.
This is motivated by the fact that the regression outputs can better follow actors and adapt to action tubes than the input proposals.
Temporal extension focuses on improving classification accuracy by incorporating longer-range temporal information.
However, simply taking a longer clip as input is inefficient and also ineffective since a longer sequence tends to have larger spatial displacement, as shown in Figure~\ref{fig:1}.
Instead, we progressively process longer sequences at each step and adaptively extend proposals to follow action movement.
In this manner, STEP can naturally handle the spatial displacement problem and therefore provide more efficient and effective spatio-temporal modeling.
Moreover, STEP achieves superior performance by using only a handful (e.g., 11) of proposals, obviating the need to generate and process large numbers (e.g., $>$1K) of proposals due to the tremendous spatial and temporal search space.

To our knowledge, this work provides the first end-to-end progressive optimization framework for video action detection. We bring up the spatial displacement problem in action tubes and show that our method can naturally handle the problem in an efficient and effective way. Extensive evaluations find our approach to produce superior detection results while only using a small number of proposals.  

\section{Related Work}


\textbf{Action Recognition.} A large family of the research in video action recognition is about action classification, which provides fundamental tools for action detection, such as two-stream networks on multiple modalities~\cite{simonyan2014two, multilayer}, 3D-CNN for simultaneous spatial and temporal feature learning~\cite{carreira2017quo, res3d}, and RNNs to capture temporal context and handle variable-length video sequences~\cite{beyond-snippets, prernn}. 
Another active research line is the temporal action detection, which focuses on localizing the temporal extent of each action. Many methods have been proposed, from fast temporal action proposals~\cite{fast-temporal-proposal}, region convolutional 3D network~\cite{r-c3d}, to budget-aware recurrent policy network~\cite{budget-aware}.  

\textbf{Spatio-Temporal Action Detection.} 
Inspired by the recent advances in image object detection, a number of efforts have been made to extend image object detectors (e.g., R-CNN, Fast R-CNN and SSD) to the task as frame-level action detectors~\cite{gkioxari2015finding, peng2016multi, saha2016deep, singh2017online, weinzaepfel2015learning, yang2017spatio, r3dcnn}.
The extensions mainly include: first, optical flow is used to capture motion cues, and second, linking algorithms are developed to connect frame-level detection results as action tubes.  
Although these methods have achieved promising results, the temporal property of videos is not explicitly or fully exploited as the detection is performed on each frame independently.
To better leverage the temporal cues, several recent work has been proposed to perform action detection at clip level.
For instance, ACT \cite{kalogeiton2017action} takes as input a short sequence of frames (e.g., 6 frames) and outputs the regressed tubelets, which are then linked by a tubelet linking algorithm to construct action tubes.
Gu et al. \cite{gu2017ava} further demonstrate the importance of temporal information by using longer clips (e.g., 40 frames) and taking advantage of I3D pre-trained on the large-scale video dataset~\cite{carreira2017quo}.
Rather than linking the frame or clip level detection results, there are also some methods that are developed to link the proposals before classification to generate action tube proposals~\cite{hou2017tube,li2018recurrent}.


\textbf{Progressive Optimization.} This technique has been explored in a range of vision tasks from pose estimation \cite{carreira2016human}, image generation \cite{gregor2015draw} to object detection \cite{ cai2017cascade,gidaris2015object,gidaris2016attend,najibi2016g}. 
Specifically, the multi-region detector \cite{gidaris2015object} introduces iterative bounding box regression with R-CNN to produce better regression results.
AttractioNet in \cite{gidaris2016attend} employs a multi-stage procedure to generate accurate object proposals that are then input to Fast R-CNN.
G-CNN \cite{najibi2016g} trains a regressor to iteratively move a grid of bounding boxes towards objects.
Cascade R-CNN \cite{cai2017cascade} proposes a cascade framework for high-quality object detection, where a sequence of R-CNN detectors are trained with increasing IoU thresholds to iteratively suppress close false positives.

\section{Method}
\label{sec:approach}

In this section, we introduce the proposed progressive learning framework STEP for video action detection.
We first formulate the problem and provide an overview of our approach.
We then describe in details the two primary components of STEP including spatial refinement and temporal extension.
Finally, the training algorithm and implementation details are presented.

\begin{figure}
\begin{center}
\includegraphics[width=\columnwidth]{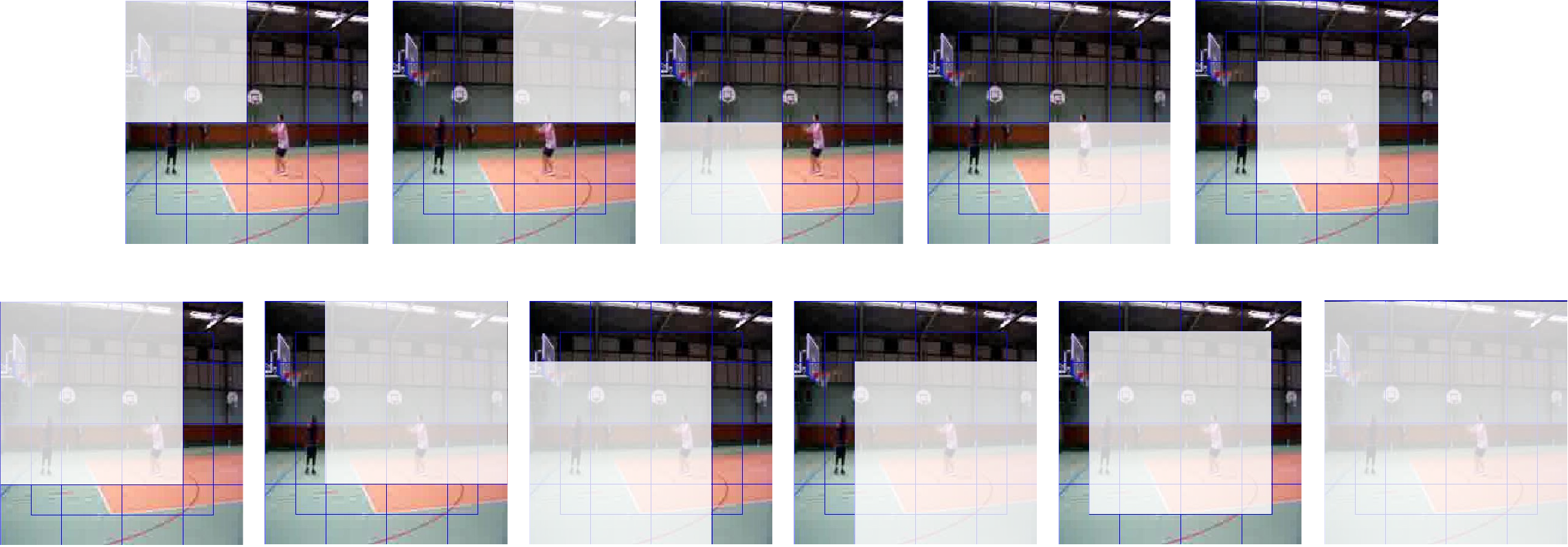}
\end{center}
\vspace{-.1in}
\caption{Example of the 11 initial proposals: 2D boxes are replicated across time to obtain cuboids.}
\label{fig:2}
\end{figure}

\subsection{Framework Overview}
\label{sec:overview}
Proceeding with the recent work \cite{gu2017ava,kalogeiton2017action}, our approach performs action detection at clip level, i.e., detection results are first obtained from each clip and then linked to build action tubes across a whole video.
We assume that each action tubelet of a clip has a constant action label, considering the short duration of a clip, e.g., within one second. 

Our target is to tackle the action detection problem through a few progressive steps, rather than directly detecting actions all at one run.
In order to detect the actions in a clip $I_t$ with $K$ frames, according to the maximum progressive steps $S_{max}$, we first extract the convolutional features for a set of clips $\mathcal{I} = \left \{I_{t-S_{max}+1},...,I_{t},...,I_{t+S_{max}-1}\right \}$ using a backbone network such as VGG16~\cite{simonyan2014very} or I3D~\cite{carreira2017quo}.
The progressive learning starts with $M$ pre-defined proposal cuboids $B^0 = \left\{b^0_i\right\}_{i = 1}^M$ and $b^0_i\in \mathcal{R}^{K\times4}$, which are sparsely sampled from a coarse-scale grid of boxes and replicated across time to form the initial proposals.
An example of the 11 initial proposals used in our experiments is illustrated in Figure \ref{fig:2}.
These initial proposals are then progressively updated to better classify and localize the actions.
At each step $s$, 
we update the proposals by performing the following processes in order:

\begin{itemize}[leftmargin=*,noitemsep,topsep=0pt]
	\item \textbf{Extend}: the proposals are temporally extended to the adjacent clips to include longer-range temporal context, 
    and the temporal extension is adaptive to the movement of actions, as described in Section~\ref{sec:temporal}.
	\item \textbf{Refine}: the extended proposals are forwarded to the spatial refinement, which outputs the classification and regression results, as presented in Section~\ref{sec:spatial}.
	\item \textbf{Update}: all proposals are updated using a simple greedy algorithm, i.e., each proposal is replaced by the regression output with the highest classification score:
\end{itemize}

\begin{equation}
\label{eq:1}
b^s_i \doteq l_i^s(c^*), \ \  c^* = \argmax_{c} p_i^s(c),
\end{equation}
where $c$ is an action class, $p_i^s \in \mathcal{R}^{(C+1)}$ is the probability distribution of the $i$th proposal over $C$ action classes plus background, $l_i^s \in \mathcal{R}^{K \times 4 \times C}$ denotes its parameterized coordinates (for computing the localization loss in Eq.~\ref{eq:3}) at each frame for each class, and $\doteq$ indicates decoding the parameterized coordinates. 
We summarize the outline of our detection algorithm in Algorithm \ref{alg:1}. 

\begin{algorithm}[t]
\SetKwFunction{Extend}{Extend }\SetKwFunction{Refine}{Refine }\SetKwFunction{Update}{Update }
\SetKwInOut{Input}{Input}\SetKwInOut{Output}{Output}
\Input{video clips $\mathcal{I}$, initial proposals $B^0$, and maximum steps $S_{max}$}
\Output{detection results $\big\{(p_i^{S_{max}}, l_i^{S_{max}})\big\}_{i=1}^M$}
extract convolutional features for video clips $\mathcal{I}$ \\
\For{$s\leftarrow 1$ \KwTo $S_{max}$}{
    \eIf{$s == 1$}{
        \tcp{initial proposals}
        $\tilde{B}^{s-1} \leftarrow B^0$\ 
    }
    {
        \tcp{temporal extension (Sec.\ref{sec:temporal})}
        $\tilde{B}^{s-1} \leftarrow$ Extend\hspace{0.3mm}$(B^{s-1})$\\ 
    }
    \tcp{spatial refinement (Sec.\ref{sec:spatial})}
    $\big\{(p_i^s, l_i^s)\big\}_{i=1}^M \leftarrow $ Refine\hspace{0.3mm}$(\tilde{B}^{s-1})$\\ 
    \tcp{update proposals (Eq.\ref{eq:1})}
    $B^s \leftarrow $ Update\hspace{0.3mm}$\big(\big\{(p_i^s, l_i^s)\big\}_{i=1}^M\big)$\\ 
}
\caption{STEP Action Detection for Clip $I_t$}\label{alg:1}
\end{algorithm}

\subsection{Spatial Refinement}
\label{sec:spatial}
\begin{figure*}
\begin{center}
\includegraphics[width=0.95\textwidth]{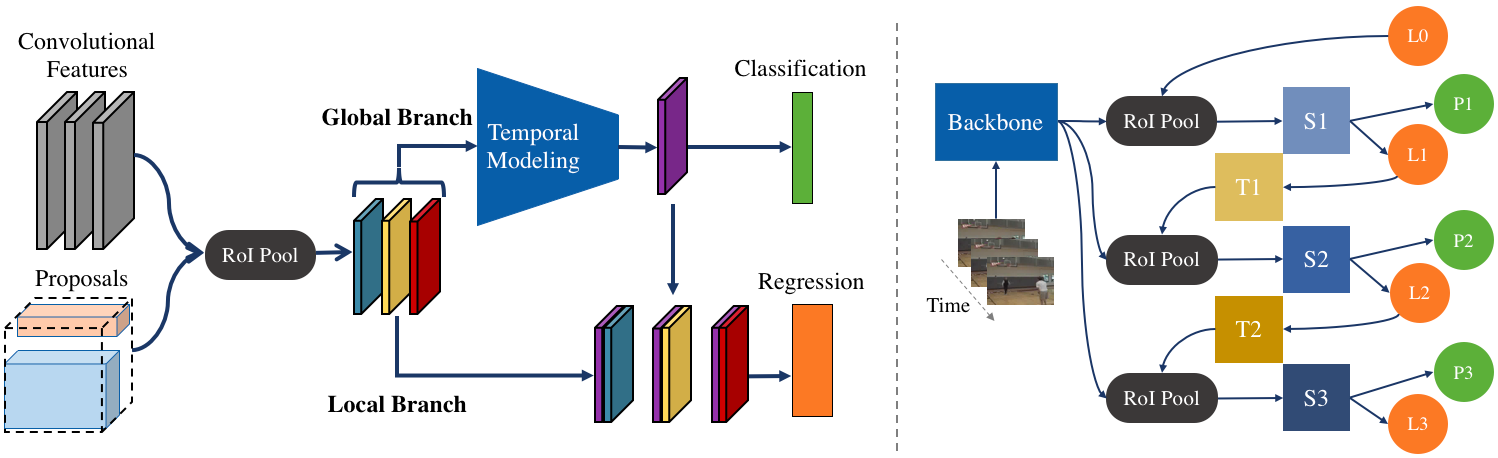}
\end{center}
\vspace{-.1in}
\caption{Left: the architecture of our two-branch network. Right: the illustration of our progressive learning framework, where ``S" indicates spatial refinement, ``T" temporal extension, ``P" classification, and ``L" localization, the numbers correspond to the steps, and ``L0" denotes the initial proposals.}
\label{fig:architecture}
\end{figure*}

At each step $s$, the spatial refinement solves a multi-task learning problem that involves action classification and localization regression.
Accordingly, we design a two-branch architecture, which learns separate features for the two tasks, as illustrated in Figure~\ref{fig:architecture}.
Our motivation is that the two tasks have substantially different objectives and require different types of information.
For accurate action classification, it demands context features in both space and time, while for robust localization regression, it needs more precise spatial cues at frame level.
As a result, our two-branch network consists of a \textit{global branch} that performs spatio-temporal modeling on the entire input sequence for action classification, as well as a \textit{local branch} that performs bounding box regression at each frame.

Given the frame-level convolutional features and the tubelet proposals for the current step, we first extract regional features through an ROI pooling~\cite{girshick2015fast}.
Then we take the regional features to the global branch for spatio-temporal modeling and produce the global feature.
Each global feature encodes the context information of a whole tubelet and is further used to predict the classification output $p_i^s$.
Moreover, the global feature is concatenated with the corresponding regional features at each frame to form the local feature, which is used to generate the class-specific regression output $l_i^s$.
Our local feature not only captures the spatio-temporal context of a tubelet but also extracts the local details of each frame.
By jointly training the two branches, the network learns the two separate features that are informative and adaptable for their own tasks. 

\textbf{Training Loss.}
We enforce a multi-task loss to jointly train for action classification and tubelet regression.
Let $\mathcal{P}^s$ denote the set of selected positive samples and $\mathcal{N}^s$ the set of negative samples at step $s$ (the sampling strategy is described in Section \ref{sec:optimization}).
We define the training loss $\mathcal{L}^s$ as:
\begin{equation}
\label{eq:2}
    \mathcal{L}^s = \hspace{-3.5mm} \sum_{i\in \left\{\mathcal{P}^s, \hspace{0.5mm} \mathcal{N}^s\right\}}\hspace{-4mm}\mathcal{L}_{cls} (p_i^s, u_i) + \lambda \sum_{i\in \mathcal{P}^s} \mathcal{L}_{loc}(l_i^s(u_i), v_i),
\end{equation}
where $u_i$ and $v_i$ are the ground truth class label and localization target for the $i$th sample, and $\lambda$ is the weight to control the importance of the two loss terms. 
We employ the multi-class cross-entropy loss as the classification loss $\mathcal{L}_{cls}(p_i^s, u_i) = -\log p_i^s(u_i)$ in Eq.~\ref{eq:2}. 
We define the localization loss using the averaged $\ell_{1, \text{smooth}}$ between predicted and ground truth bounding boxes over the frames of a clip:
\begin{equation}
\label{eq:3}
    \mathcal{L}_{loc}(l_i^s(u_i), v_i) = \frac{1}{K} \sum_{
k=1}^K \ell_{1, \text{smooth}}(l_{i, k}^s(u_i) - v_{i, k}).
\end{equation}
We apply the same parameterization for $v_{i, k}$ as in \cite{girshick2014rich} by using a scale-invariant center translation and a log-space height/width shift relative to the bounding box.

\subsection{Temporal Extension}
\label{sec:temporal}
Video temporal information, especially the long-term temporal dependency, is critical for accurate action classification \cite{carreira2017quo,prernn}.
In order to leverage longer range of temporal context, we extend the proposals to include in more frames as input.
However, the extension is not trivial since the spatial displacement problem becomes even more severe for longer sequences, as illustrated in Figure \ref{fig:1}.
Recently, some negative impacts caused by the spatial displacement problem for action detection have also been observed by \cite{gu2017ava,kalogeiton2017action}, which simply replicate 2D proposals across time to increase longer temporal length.

With the intention to alleviate the spatial displacement problem, we perform temporal extension \textit{progressively} and \textit{adaptively}.
From the second step, we extend the tubelet proposals to the two adjacent clips at a time.
In other words, at each step $1 \leq s < S_{max}$, the proposals $B^{s}$ with length $K^{s}$ are extended to $\tilde{B}^{s} = B^{s}_{-1} \circ B^{s} \circ B^{s}_{+1}$ with length $K^{s} + 2K$, where $\circ$ denotes concatenation.
Additionally, the temporal extension is adaptive to action movement by taking advantage of the regressed tubelets from the previous step.
We introduce two methods to enable the temporal extension to be adaptive as described in the following.

\textbf{Extrapolation.} By assuming that the spatial movement of an action satisfies a linear function approximately within a short temporal range, such as a 6-frame clip, we can extend the tubelet proposals by using a simple linear extrapolation function:
\begin{align}
B_{+1,k}^{s} = B_{K^{s}}^{s} + \frac{k}{K-1}(B_{K^{s}}^{s} - B_{K^{s}-K+1}^{s}).
\end{align}
A similar function can be applied to $B_{-1}^{s}$ to adapt to the movement trend, but the assumption would be violated for long sequences and therefore results in drifted estimations.

\textbf{Anticipation.} We can also achieve the adaptive temporal extension by location anticipation, i.e., training an extra regression branch to conjecture the tubelet locations in adjacent clips based on the current clip.
Intuitively, the anticipation requires the network to infer the movement trend in adjacent clips by action modeling in the current clip.
A similar idea is explored in \cite{yang2017spatio}, where location anticipation is used at the region proposal stage.

We formulate our location anticipation as a residual learning problem \cite{he2016deep,long2016unsupervised} based on the assumption that the tubelets of two adjacent clips differ from each other by a small residual.
Let $x$ indicate the features forwarded to the output layer $f$ of the location regressor $L^{s} = f(x)$ at step $s$.
So the anticipated locations can be obtained as:
\begin{equation}
    L_{-1}^{s} = L^{s} + f_{-1}(x), \ \ L_{+1}^{s} = L^{s} + f_{+1}(x),
\end{equation}
where $f_{-1}$ and $f_{+1}$ are the anticipation regressors, which are lightweight and introduce negligible computational overhead. $L_{-1}^{s}$ and $L_{+1}^{s}$ are then decoded to the proposals $B_{-1}^{s}$ and $B_{+1}^{s}$.
The loss function of location anticipation is defined in a similar way as Eq.~\ref{eq:3}, and combined with $\mathcal{L}_{cls}$ and $\mathcal{L}_{loc}$ with a coefficient $\gamma$ to form the overall loss.

\subsection{Network Training}
\label{sec:optimization}
Although STEP involves multiple progressive steps, the whole framework can be trained end-to-end to optimize the models at different steps jointly. 
Compared against the step-wised training scheme used in \cite{najibi2016g}, our joint training is simpler to implement, runs more efficiently, and achieves better performance in our experiments.

Given a mini-batch of training data, we first perform an ($S_{max}-1$)-step inference pass, as illustrated in the right of Figure~\ref{fig:architecture}, to obtain the inputs needed for all progressive steps.  
In practice, the detection outputs $\left\{(p_i^s, l_i^s)\right\}_{i = 1}^M$ at each step are collected and used to select the positive and negative samples $\mathcal{P}^s$ and $\mathcal{N}^s$ for training. 
We accumulate the losses of all steps and back-propagate to update the whole model at the same time.

\begin{figure}[t]
\begin{center}
\includegraphics[width=\columnwidth]{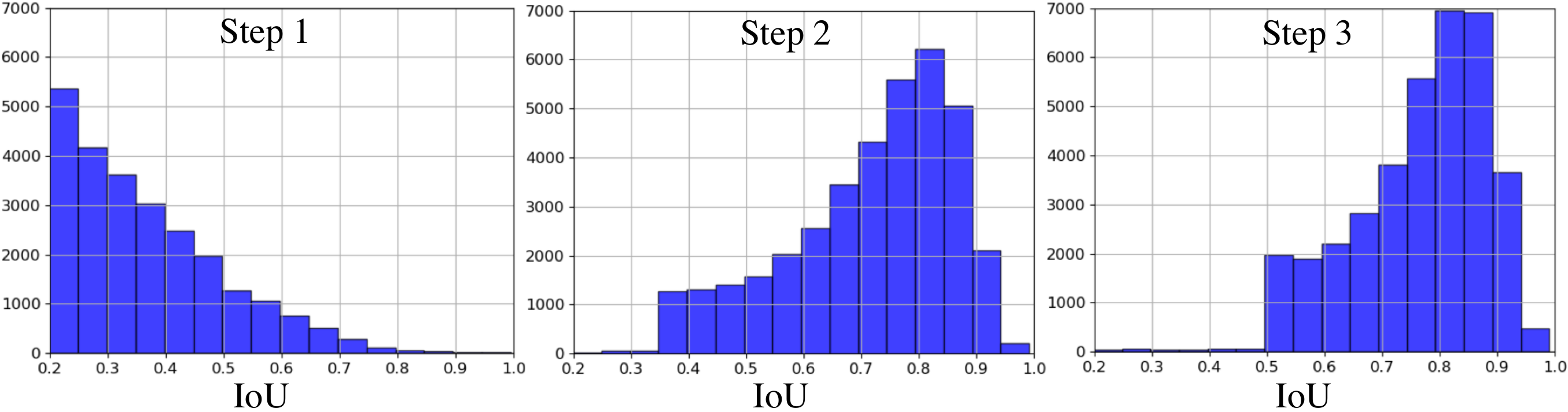}
\end{center}\vspace{-.2in}
\caption{Change of input distribution (IoU between input proposals and ground truth) over steps on UCF101.}
\label{fig:distribution}
\end{figure}

\textbf{Distribution Change.} Compared to the prior work that performs detection in one run, our training could be more challenging as the input/output distributions change over steps.
As shown in Figure~\ref{fig:distribution}, the input distribution is right-skewed or centered in a low-IoU level at early steps, and reverses at later steps.
This is because our approach starts from a coarse-scale grid (see Figure~\ref{fig:2}) and progressively refines them towards generating high-quality proposals.
Accordingly, the range of output distribution (i.e., the scale of offset vectors) decreases over steps.

Inspired by \cite{cai2017cascade}, we tackle the distribution change in three ways. 
First, separate headers are used at different steps to adapt to the different input/output distributions. 
Second, we increase IoU thresholds over the multiple steps.
Intuitively, a lower IoU threshold at early steps tolerates the initial proposals to include sufficient positive samples and a higher IoU threshold at late steps encourages high-quality detection.
Third, a hard-aware sampling strategy is employed to select more informative samples during training. 

\textbf{Hard-Aware Sampling.} We design the sampling strategy based on two principles: (\romannumeral 1) the numbers of positive and negative samples should be roughly balanced, and (\romannumeral 2) the harder negatives should be selected more often.
To measure the ``hardness" of a negative sample, we use the classification scores from the previous step. The tubelet with a high confidence but a low overlap to any ground truth is viewed as a hard sample.
We calculate the overlap of two tubelets by averaging the IoU of bounding boxes over $K$ frames of the target clip.
So the negative samples with higher classification scores will be sampled with a higher chance.

Formally, given a set of proposals and the overlap threshold $\tau^s$ at step $s$, we first assign positive labels to the candidates with the highest overlap with ground truth.
This is to ensure that each ground truth tube has at least one positive sample.
After that, the proposals having an overlap higher than $\tau^s$ with any ground truth tube are added to the positive pool and the rest to the negative pool.
We then sample $|\mathcal{P}^s|$ positives and $|\mathcal{N}^s|$ negatives from the two pools, respectively, with the sampling probability proportional to the classification score.
For the first step, the highest overlap with ground truth tubes is used as the score for sampling.
Each selected positive in $\mathcal{P}^s$ is assigned to the ground truth tube with which it has the highest overlap.
Note that a single proposal can be assigned to only one ground truth tube.

\subsection{Full Model}
\label{sec:full}
We can also integrate our model with the common practices for video action detection~\cite{gu2017ava,kalogeiton2017action,singh2017online}, such as two-stream fusion and tubelet linking.

\textbf{Scene Context.}
It has been proven to be beneficial to object and action detection \cite{li2018recurrent,sun2018actor}.
Intuitively, some action-related semantic clues from scene context can be utilized to improve action classification, for example, the scene of a basketball court for recognizing ``basketball dunk".
We incorporate scene context by concatenating extended features to original regional features in the global branch. 
The extended features can be obtained by RoI pooling of the whole image.
So the global features encode both spatial and temporal context useful for action classification.

\textbf{Two-Stream Fusion.} 
Most previous methods use late fusion to combine the results at test time, i.e., the detections are obtained independently from the two streams and then fused using either mean fusion \cite{kalogeiton2017action} or union fusion \cite{singh2017online}.
In this work, we also investigate early fusion for two-stream fusion, which concatenates RGB frames and optical flow maps in channel and input to the network as a whole.
Intuitively, early fusion can model the low-level interactions between the two modalities and also obviates the need for training two separate networks.
In addition, a hybrid fusion can be further performed to combine detection results from the early fusion and the two streams.
Our experiment shows that early fusion outperforms late fusion, and hybrid fusion achieves the best performance.

\textbf{Tubelet Linking.} Given the clip-level detection results,  we link them in space and time to construct the final action tubes.
We follow the same linking algorithm as described in \cite{kalogeiton2017action}, apart from that we do not apply global non-maximum suppression across classes but perform temporal trimming over the linked paths as commonly used in \cite{li2018recurrent,saha2016deep}.
The temporal trimming enforces consecutive boxes to have smooth classification scores by solving an energy maximization problem via dynamic programming.

\section{Experiments}

In this section, we describe the experiments to evaluate STEP and compare against the recent competing algorithms.
We start by performing a variety of ablation studies to better understand the contributions of each individual component in our approach.
We then report comparisons to the state-of-the-art methods, provide in-depth analysis, and present the qualitative detection results.

\subsection{Experimental Setup}
\textbf{Datasets.}
We evaluate our approach on the two benchmarks: UCF101~\cite{soomro2012ucf101} and AVA~\cite{gu2017ava}. In comparison with other action detection datasets, such as J-HMDB and UCFSports, the two benchmarks are much larger and more challenging, and more importantly, they are temporally untrimmed, which fits better to the spatio-temporal action detection task. 
UCF101 is originally an action classification dataset collected from online videos, and a subset of 24 classes with 3,207 videos are provided with the spatio-temporal annotations for action detection.
Following the standard evaluation protocol \cite{kalogeiton2017action}, we report results on the first split of the dataset.
AVA contains complex actions and scenes sourced from movies.
We use the version 2.1 of AVA, which consists of the annotations at 1 fps over 80 action classes.
Following the standard setup in \cite{gu2017ava}, we report results on the most frequent 60 classes that have at least 25 validation examples per class.

\textbf{Evaluation Metrics.}
We report the frame-level mean average precision (frame-mAP) with an IoU threshold of 0.5 for both datasets.
This metric allows us to evaluate the quality of the detection results independently of the linking algorithm.
We also use the video-mAP on UCF101 to compare with the state-of-the-art results.

\textbf{Implementation Details.}
For the experiments on UCF101, we use VGG16 \cite{simonyan2014very} pre-trained on ImageNet~\cite{imagenet} as the backbone network.
Although more advanced models are available, we choose the same backbone as \cite{kalogeiton2017action} for fair comparisons.
For the temporal modeling in global branch, we use three 3D convolutional layers with adaptive max pooling along the temporal dimension.
All frames are resized to $400\times400$ and the clip length is set to $K=6$.
Similar to \cite{kalogeiton2017action}, 5 consecutive optical flow maps are stacked as a whole for the optical flow input.
We train our models for 35 epochs using Adam \cite{kingma2014adam} with a batch size of 4.
We set the initial learning rate to $5 \times 10^{-5}$ and perform step decay after 20 and 30 epochs with the decay rate 0.1.

For the experiments on AVA, we adopt I3D \cite{carreira2017quo} (up to \texttt{Mixed\_4f}) pre-trained on Kinetics-400 \cite{kay2017kinetics} as the backbone network.
We take the two layers \texttt{Mixed\_5b} and \texttt{Mixed\_5c} of I3D for temporal modeling in our global branch.
All frames are resized to $400\times400$ and the clip length is set to $K=12$.
We use 34 initial proposals and perform temporal extension only at the third step.
As the classification is more challenging on AVA, we first pre-train our model for an action classification task using the spatial ground truth of training set.
We then train the model for action detection with a batch size of 4 for 10 epochs. We do not use optical flow on this dataset due to the heavy computation and instead combine results of two RGB models. 
Our initial learning rate is $5\times10^{-6}$ for the backbone network and $5\times10^{-5}$ for the two-branch networks, and step decay is performed after 6 epochs with the decay rate 0.1.

For all experiments, we extract optical flow (if used) with Brox \cite{brox2004high}, and perform data augmentation to the whole sequence of frames during training, including random flipping and cropping.
More architecture and implementation details are available in the appendix.

\begin{table}[t]
\parbox{.65\linewidth}{
\centering
\begin{tabular}{|c|c|c|c|c|}
\hline
\multirow{2}{*}{$S_{max}$} & \multicolumn{4}{c|}{$s$} \\
\cline{2-5}
& 1 & 2 & 3 & 4 \\
\hline\hline
1 & 51.5 & - & - & - \\
2 & 56.6 & 60.7 & - & - \\
3 & 57.1 & 61.8 & 62.6 & - \\
4 & \textbf{58.2} & \textbf{62.1} & \textbf{62.8} & \textbf{62.7} \\
\hline
\end{tabular}
}
\parbox{.3\linewidth}{
\centering
\begin{tabular}{|l| c|}
\hline
Mode & f-mAP  \\
\hline\hline
RGB  & 66.7 \\
Flow & 63.5 \\
\hline
Late & 70.7 \\
Early & 74.3 \\
Hybrid  & \textbf{75.0} \\
\hline
\end{tabular}
}
\caption{Comparisons of frame-mAP ($\%$) of our models trained with different numbers of steps (left), and different input modalities and fusion methods (right).}  
\end{table}

\subsection{Ablation Study}
\label{sec:ablation}
We perform various ablation experiments on UCF101 to evaluate the impacts of different design choices in our framework.
For all experiments in this section, we employ the 11 initial proposals as shown in Figure \ref{fig:2} and RGB only, unless explicitly mentioned otherwise, and frame-mAP is used as the evaluation metric.

\textbf{Effectiveness of Spatial Refinement.}
Our primary design of STEP is to progressively tackle the action detection problem through a few steps.
We thus first verify the effectiveness of progressive learning by comparing the detection results at different steps with the spatial refinement.
No temporal extension is applied in this comparison.
Table 1(a) demonstrates the step-wise performance under different maximum steps $S_{max}$.
Since our approach starts from the coarse-scale proposals, performing spatial refinement once is insufficient to achieve good results.
We observe that the second step improves results consistently and substantially, indicating that the updated proposals have higher quality and provide more precise information for classification and localization.
Further improvement can be obtained by additional steps, suggesting the effectiveness of our progressive spatial refinement.
We use 3 steps for most of our experiments as the performance saturates after that.
Note that using more steps also improves the results of early steps, due to the benefits of our multi-step joint training.

\begin{figure}[t]
\begin{center}
\includegraphics[width=0.92\columnwidth]{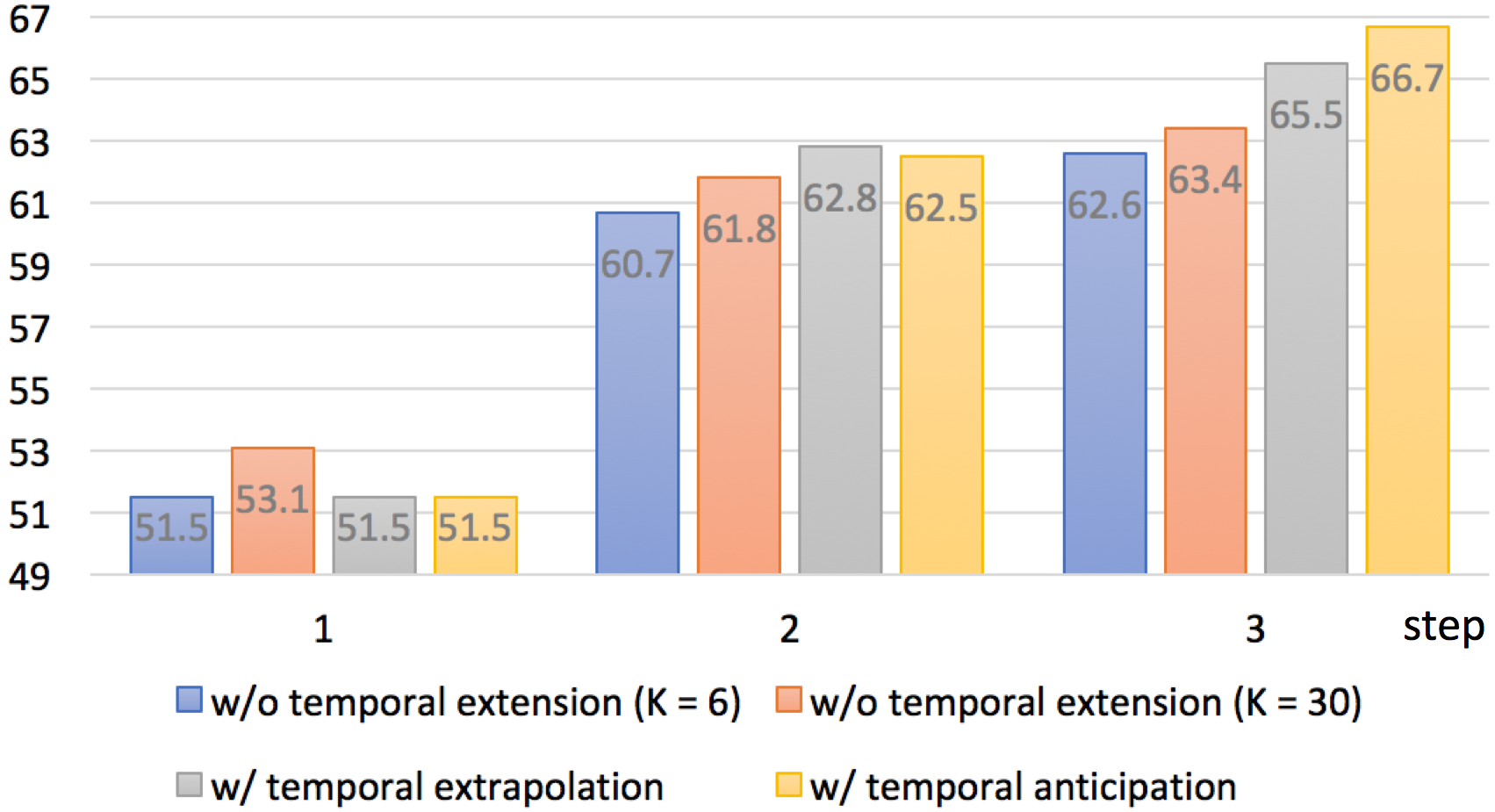}
\end{center}
\vspace{-.1in}
\caption{Comparison of frame-mAP ($\%$) of our models trained with and without temporal extension.}
\label{fig:4}
\end{figure}

\textbf{Effectiveness of Temporal Extension.}
In addition to the spatial refinement, our progressive learning contains the temporal extension to progressively process a longer sequence at each step.
We compare the detection results with and without temporal extension in Figure \ref{fig:4}.
We show the results of the models taking $K=6$ and $K=30$ frames as inputs directly without temporal extension, and the results of the extrapolation and anticipation methods.
Note that the models with temporal extension also deal with 30 frames at the third step (extension process: 6 $\rightarrow$ 18 $\rightarrow$ 30).

Both of the temporal extension methods outperform the baseline ($K = 6$) by a large margin, which clearly shows the benefit of incorporating longer-range temporal context for action classification.
More remarkably, simply taking $K = 30$ frames as input without temporal extension results in inferior performance, validating the importance of adaptively extending the temporal scale in the progressive manner.
Furthermore, we observe that anticipation performs better than extrapolation for longer sequences, indicating that anticipation can better capture nonlinear movement trends and therefore generate better extensions.

\textbf{Fusion Comparison.}
Table 1(b) presents the detection results of different fusions: late, early and hybrid fusion.
In all cases, using both modalities improves the performance compared to individual ones.
We find that early fusion outperforms late fusion, and attribute the improvement to modeling between the two modalities at the early stage.
Hybrid fusion achieves the best result by further utilizing the complementary information of different methods.

\textbf{Miscellaneous.}
We describe several techniques to improve the training in Section~\ref{sec:approach}, including incorporating scene context, hard-award sampling and increasing IoU threshold.
To validate the contributions of the three techniques, we conduct ablation experiments by removing one at a time, which correspondingly results in a performance drop of 2.5\%, 1.5\% and 1\%.
In addition, we observe that incorporating scene context provides more gains for later steps, suggesting that scene context is more important for action classification when bounding boxes become tight.

\begin{figure}[t]
\begin{center}
\includegraphics[width=\columnwidth]{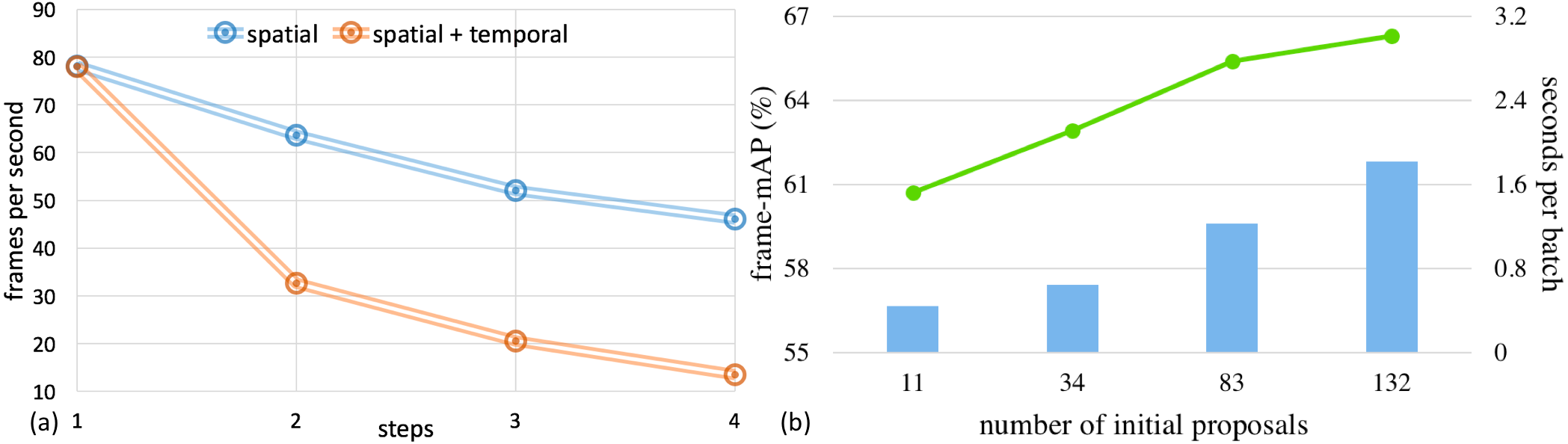}
\end{center}
\vspace{-.1in}
\caption{Analysis of runtime of our approach under various settings: (a) the inference speeds using different step numbers with and without temporal extension, and (b) the detection results (green dots) and speeds (blue bars) using different numbers of initial proposals.}
\label{fig:runtime}
\end{figure}

\subsection{Runtime Analysis}
Although STEP involves a multi-step optimization, our model is efficient since we only process a small number of proposals. STEP runs at 21 fps using early fusion with 11 initial proposals and 3 steps on a single GPU, which is comparable with the clip based approach (23 fps)~\cite{kalogeiton2017action} and much faster than the frame based method (4 fps)~\cite{peng2016multi}. Figure~\ref{fig:runtime}(a) demonstrates the speeds of our approach with increasing number of steps under the settings with and without temporal extension. We also report the running time and detection performance of our approach (w/o temporal extension for 3 steps) with increasing number of initial proposals in Figure~\ref{fig:runtime}(b). We observe substantial gains in detection accuracy by increasing the number of initial proposals, but it also results in slowed inference speed. This trade-off between accuracy and speed can be controlled according to a specified time budget.               

\subsection{Comparison with State-of-the-Art Results}
\label{sec:compare}
We compare our approach with the state-of-the-art methods on UCF101 and AVA in Tables \ref{tab:3} and \ref{tab:4}.
Following the standard settings, we report the frame-mAP at IoU threshold 0.5 on both datasets and the video-mAP at various IoU thresholds on UCF101.
STEP consistently performs better than the state-of-the-art methods on UCF101, and brings a clear gain in frame-mAP, producing $5.5\%$ improvement over the second best result. Our approach also achieves superior result on AVA, outperforming the recently proposed ACRN by $1.2\%$. Notably, STEP performs detection simply from a handful of initial proposals, while other competing algorithms rely on a great amount of densely sampled anchors or an extra person detector trained with external large-scale image object detection datasets.

\begin{figure*}[t]
\begin{center}
\includegraphics[width=0.9\textwidth]{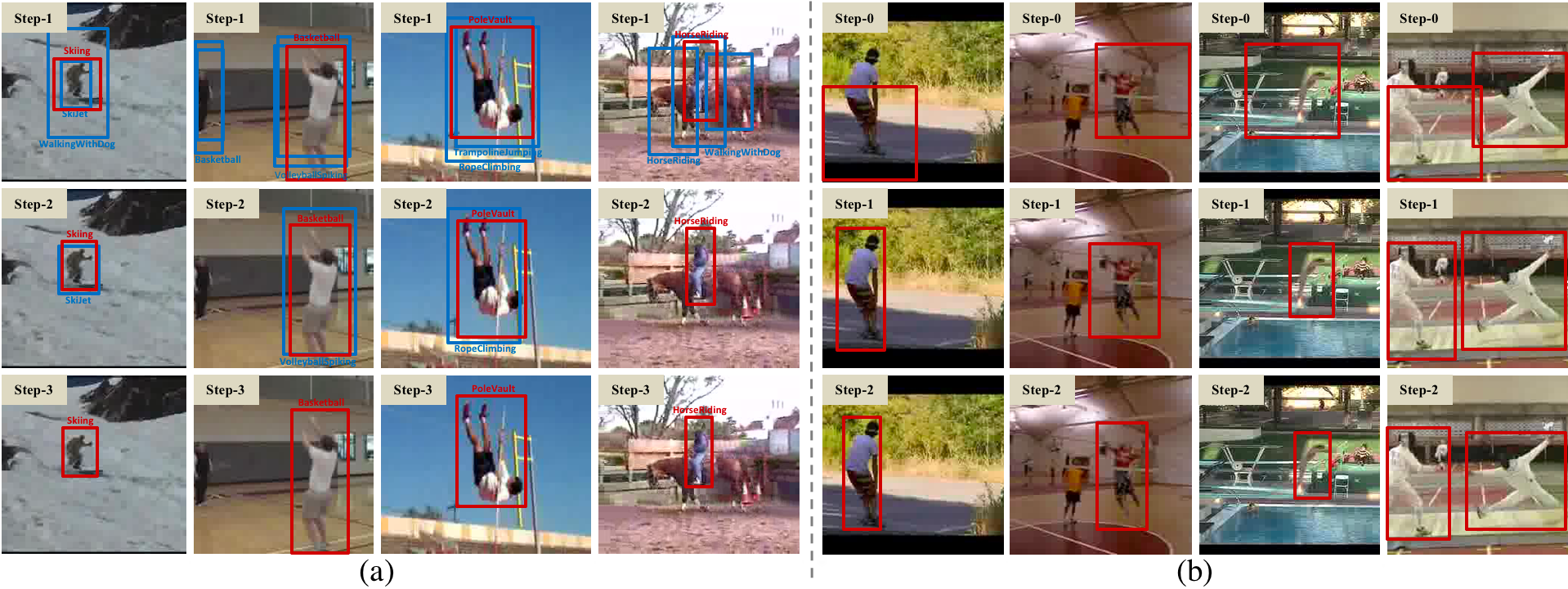}
\end{center}
\vspace{-.15in}
\caption{Examples of the detection results on UCF101. Red boxes indicate correct detection and blue ones misclassification. (a) illustrates the effect of progressive learning to improve action classification over steps. (b) demonstrates the regression outputs by spatial refinement at each step.} 
\label{fig:3}
\end{figure*}

\begin{table}
\begin{center}
\begin{tabular}{|l|c| c c c |}
\hline
\multirow{2}{*}{Method}  & frame-mAP & \multicolumn{3}{c|}{video-mAP} \\
 \cline{2-5}
 & 0.5 & 0.05 & 0.1 & 0.2 \\
\hline\hline
MR-TS \cite{peng2016multi}& 65.7 & 78.8 & 77.3 & 72.9 \\
ROAD \cite{singh2017online} & - & - & - & 73.5 \\
CPLA \cite{yang2017spatio} & - & 79.0 & 77.3 & 73.5 \\
RTPR \cite{li2018recurrent} & - & 81.5 & 80.7 & 76.3 \\
PntMatch \cite{r3dcnn} & 67.0 & 79.4 & 77.7 & 76.2 \\
T-CNN \cite{hou2017tube} & 67.3 & 78.2 & 77.9 & 73.1 \\
ACT \cite{kalogeiton2017action} & 69.5 & - & - & 76.5 \\
\hline
Ours & \textbf{75.0} & \textbf{84.6} & \textbf{83.1} & \textbf{76.6} \\
\hline
\end{tabular}
\end{center}
\vspace{-.125in}
\caption{Comparison with the state-of-the-art methods on UCF101 by frame-mAP ($\%$) and video-mAP ($\%$) under different IoU thresholds.}
\label{tab:3}
\end{table}

\begin{table}
\begin{center}
\begin{tabular}{|l|c |}
\hline
  Method & frame-mAP \\
\hline\hline
Single Frame$^*$ \cite{gu2017ava} & 14.2 \\
I3D \cite{gu2017ava} & 14.7 \\
I3D$^*$ \cite{gu2017ava} & 15.6 \\
ACRN$^*$ \cite{sun2018actor} & 17.4 \\
\hline
Ours & \textbf{18.6} \\
\hline
\end{tabular}
\end{center}
\vspace{-.125in}
\caption{Comparison with the state-of-the-art methods on AVA by frame-mAP ($\%$) under IoU $= 0.5$. ``*'' means the results obtained by incorporating optical flow.}
\label{tab:4}
\end{table}

\begin{figure}[th]
\begin{center}
\includegraphics[width=\columnwidth]{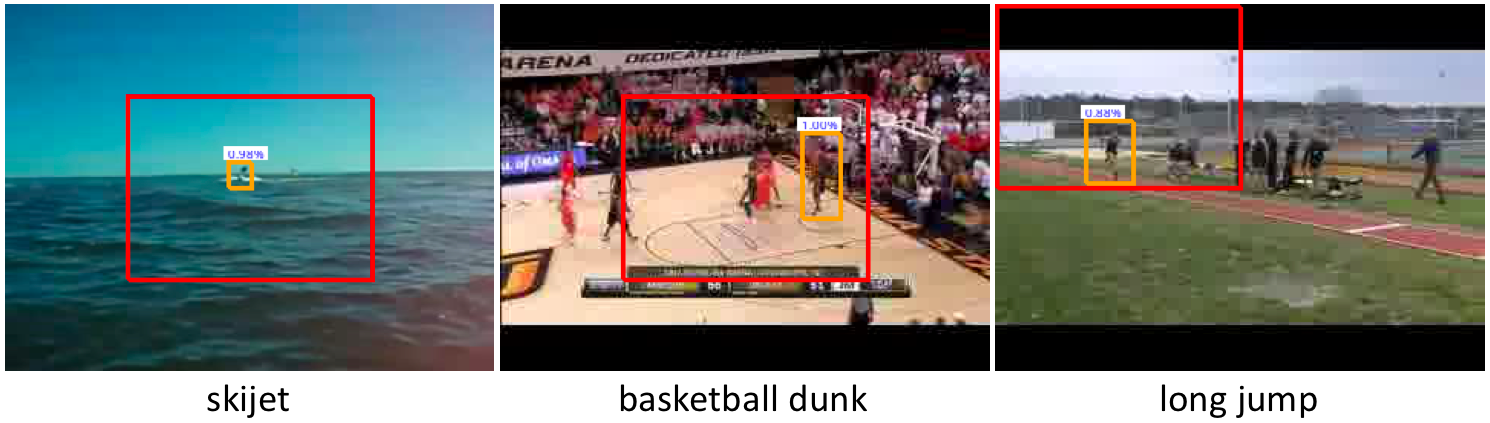}
\end{center}
\vspace{-.1in}
\caption{Examples of the small scale action detection by our approach. Red boxes indicate the initial proposals and orange ones the detection outputs.}
\label{fig:size}
\end{figure}

\subsection{Qualitative Results}
\label{sec:visualize}
We visualize the detection results of our approach at different steps in Figure \ref{fig:3}.
Each row indicates the detection outputs at a certain step.
A bounding box is labeled in red if the detection result is correct, otherwise it is labeled in blue.
Figure \ref{fig:3}(a) demonstrates the effect of progressive learning for more accurate action classification.
It can be observed by the fact that the blue boxes are eliminated at later steps.
In Figure \ref{fig:3}(b), the first row corresponds to the initial proposals and the next two rows show the effect of spatial refinement of the proposals over steps.
It is clear that the proposals progressively move towards the persons performing the actions and better localization results are obtained at later steps. Although starting from coarse-scale proposals, our approach is robust to various action scales thanks to the progressive spatial refinement, as illustrated in Figure~\ref{fig:size}.

\section{Conclusion}
In this paper, we have proposed the spatio-temporal progressive learning framework STEP for video action detection. STEP involves spatial refinement and temporal extension, where the former starts from sparse initial proposals and iteratively updates bounding boxes, and the latter gradually and adaptively increases sequence length to incorporate more related temporal context. STEP is found to be able to more effectively make use of longer temporal information by handling the spatial displacement problem in action tubes. Extensive experiments on two benchmarks show that STEP consistently brings performance gains by using only a handful of proposals and a few updating steps.         

\vspace{3mm}

\noindent\textbf{Acknowledgement.} Davis acknowledges the support from IARPA via Department of Interior/Interior Business Center (DOI/IBC) under contract number D17PC00345.  


{\small
\bibliographystyle{ieee_fullname}
\bibliography{egbib}
}

\appendix
\section*{Appendix}

In this appendix, Section \ref{sec:1} summarizes the details of our two-branch architecture and how to generate the initial proposals. Section \ref{fig:supp_spatial} presents more evidences of the spatial displacement problem in action detection. Section \ref{sec:more} provides more algorithm and result analysis.

\section{Implementation Details}
\label{sec:1}

\textbf{UCF101 Dataset.} 
Table \ref{tab:4} shows the details of our two-branch architecture.
The network takes as inputs a sequence of $512\times 25\times 25$ feature maps from the backbone network (i.e., VGG16) as well as a set of proposal tubelets.
For each proposal, an RoI pooling layer extracts a sequence of fixed-length regional features from the feature maps.
For temporal modeling in the global branch, we first spatially extend each proposal tubelet to incorporate more scene context, as described in Section 3.5 of the paper.
We then forward the extended features to three 3D convolutional layers to obtain the global features.
To perform action classification, the global features are flatten and fed into a sequence of fully connected (\texttt{fc}) layers, which finally output the softmax probability estimates over $C$ classes plus background.
To perform tubelet regression, the global features are concatenated along channel dimension with the regional features at each frame and then fed into another sequence of \texttt{fc} layers, which produce a class-specific regression output with the shape $4\times(C+1)$ for each frame.

\textbf{AVA Dataset.} 
The overall architecture is the same as the one in Table \ref{tab:4} except that we do not introduce extra 3D convolutional layers for temporal modeling.
Instead, the \texttt{Mixed\_5b} and \texttt{Mixed\_5c} in I3D are used and followed by a $1\times1\times1$ convolutional layer to downsample the channel dimension to 256.

We use 34 initial proposals in the experiments on AVA since this datasets involves more actions on average at each frame than UCF101.
We define the 34 initial proposals following the practice in~\cite{najibi2016g}. 
In details, we generate the initial proposals using a two-level spatial pyramid with [$4/3$, $2$] scales and [$5/6$, $3/4$] overlap for each spatial scale.
In other words, a sliding window with size $3W/4 \times 3H/4$ pixels and overlap ratio $5/6$ is used for the first level, and a sliding window with size $W/2 \times H/2$ pixels and overlap ratio $3/4$ is used for the second level.
Here, $W$ and $H$ denote the width and height of the frames, respectively.
We extract video frames in 12 fps and resize them to $400\times400$.

\begingroup
\setlength{\medmuskip}{0mu}
\begin{table}
\begin{center}
\begin{tabular}{c|c|c}
\multicolumn{2}{c|}{Layer} & Output size \\
\hline\hline
\multicolumn{3}{c}{Global Branch} \\ \hline
\multirow{2}{*}{conv1} & $3\times3\times3$, 1024 & \multirow{2}{*}{$6\times7\times7$} \\ 
\cline{2-2}
  & max pool &  \\ \hline
\multirow{2}{*}{conv2} & $3\times3\times3$, 512 & \multirow{2}{*}{$3\times7\times7$} \\
\cline{2-2}
 & max pool  &  \\ \hline
\multirow{2}{*}{conv3} & $3\times3\times3$, 256 & \multirow{2}{*}{$1\times7\times7$} \\
\cline{2-2}
 & average pool  &  \\ \hline
fc1(2) & 4096 & $1\times1\times1$ \\ \hline
out & $C+1$, softmax & $1\times1\times1$ \\ \hline\hline
\multicolumn{3}{c}{Local Branch} \\ \hline
fc1(2) & 4096 & $1\times1\times1$ \\ \hline
out & $4 \times (C+1)$ & $1\times1\times1$
\end{tabular}
\end{center}
\vspace{-.1in}
\caption{Architecture of the two-branch network, where $T\times H\times W, N$ represent the dimensions of convolutional kernels and output feature maps.}
\label{tab:4}
\end{table}
\endgroup

\section{Spatial Displacement}
\label{fig:supp_spatial}
\begin{figure}
\begin{center}
\includegraphics[width=\columnwidth]{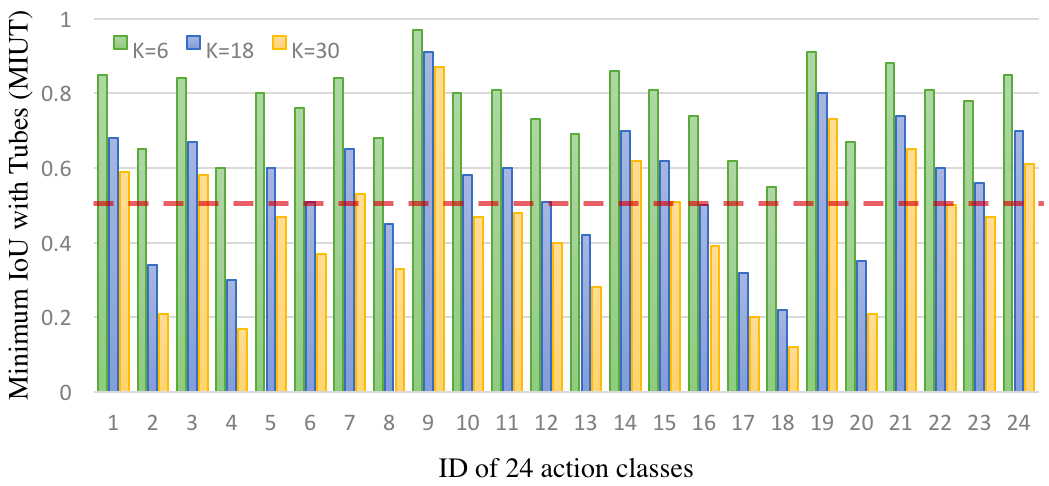}
\end{center}
\vspace{-.2in}
\caption{MIUT of ground truth action tubes on UCF101. $K$ denotes different tube lengths, and red dash line cooresponds to MIUT $= 0.5$.}
\label{fig:supp_2}
\end{figure}

\begin{figure*}[t]
\begin{center}
\includegraphics[width=\textwidth]{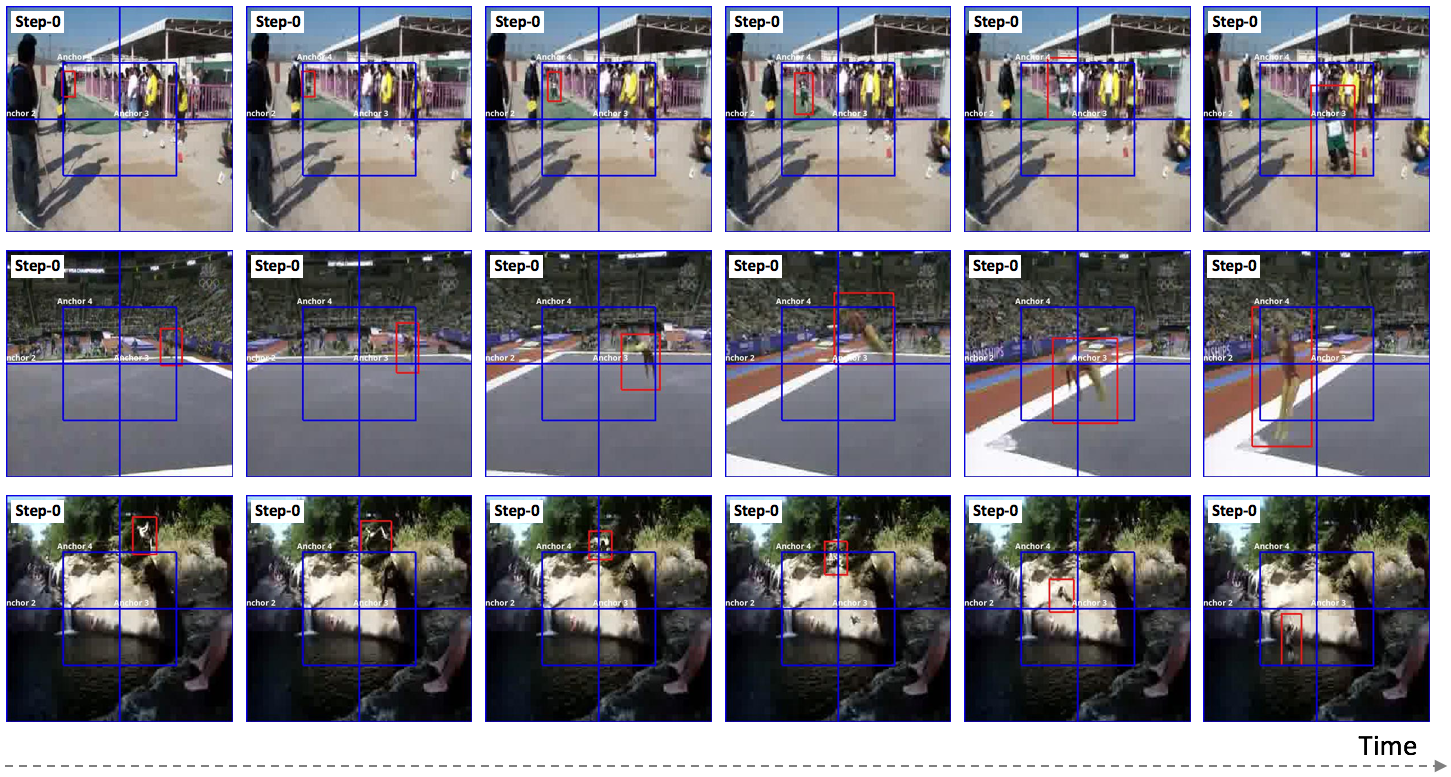}
\end{center}
\vspace{-.2in}
\caption{Examples of the spatial displacement problem. Red boxes indicate the ground truth bounding boxes and blue ones the spatial grids. From top to bottom are LongJump (ID: 12), FloorGymnastics (ID: 8) and CliffDiving (ID: 4).}
\label{fig:supp_3}
\end{figure*}

The spatial displacement problem occurs in an action tube when the sequence is long and or involves rapid movement of people or camera.
Here we analyze the spatial displacement problem on UCF101 by calculating the minimum IoU within tubes (MIUT).
Given a ground truth action tube, MIUT is defined by the minimum IoU overlap between the center bounding box (i.e., the box of the center frame) and the other bounding boxes within the tube.
Figure \ref{fig:supp_2} demonstrates the statistics of different actions with different length using ground truth action tubes in the validation set.
We observe that the spatial displacement problem is not very obvious for short clips (e.g., $K=6$), as most action classes have high MIUT values.
However, the spatial displacement problem becomes more severe for most actions when the sequence length increases.
For example, ``Skijet" (ID: 18) has a $0.12$ MIUT and ``CliffDiving" (ID: 4) has a $0.17$ MIUT when $K = 30$, indicating both actions encounter large spatial displacements within the tubes.
We also show some examples to illustrate the spatial displacement problem in Figure \ref{fig:supp_3}.

\section{More Analysis}
\label{sec:more}
\begin{figure}[t]
\begin{center}
\includegraphics[width=0.92\columnwidth]{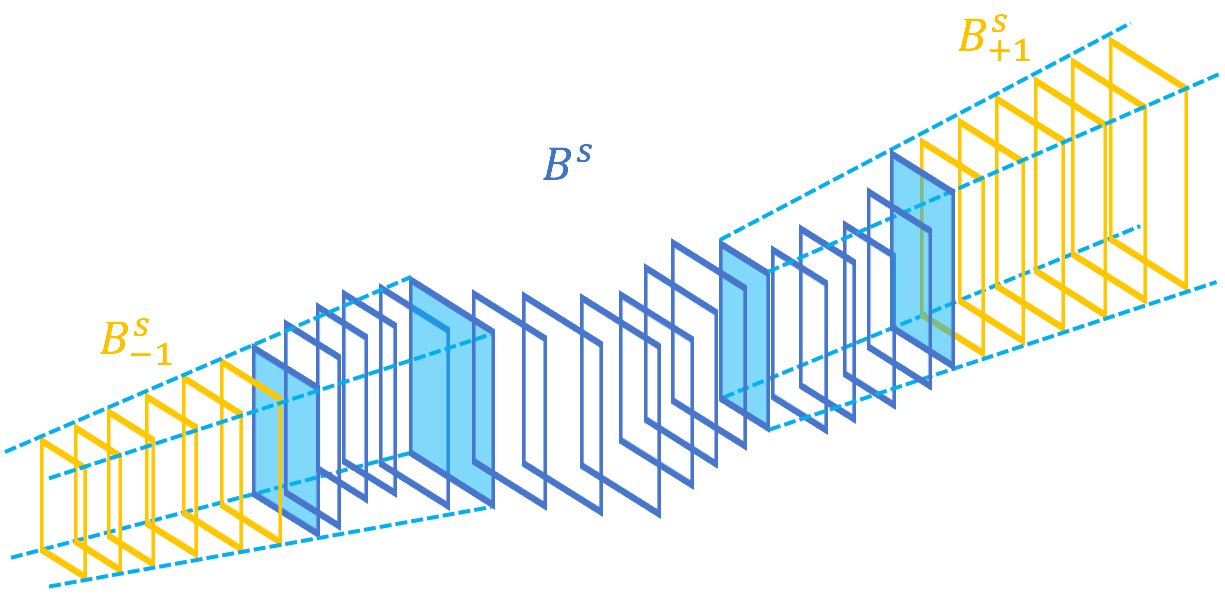}
\end{center}
\vspace{-.2in}
\caption{Illustration of the extrapolation for adaptive temporal extension. Blue shaded boxes are the first and last bounding boxes of the corresponding tubelets.}
\label{fig:extrapolate}
\end{figure}

\begin{figure}[t]
\begin{center}
\includegraphics[width=\columnwidth]{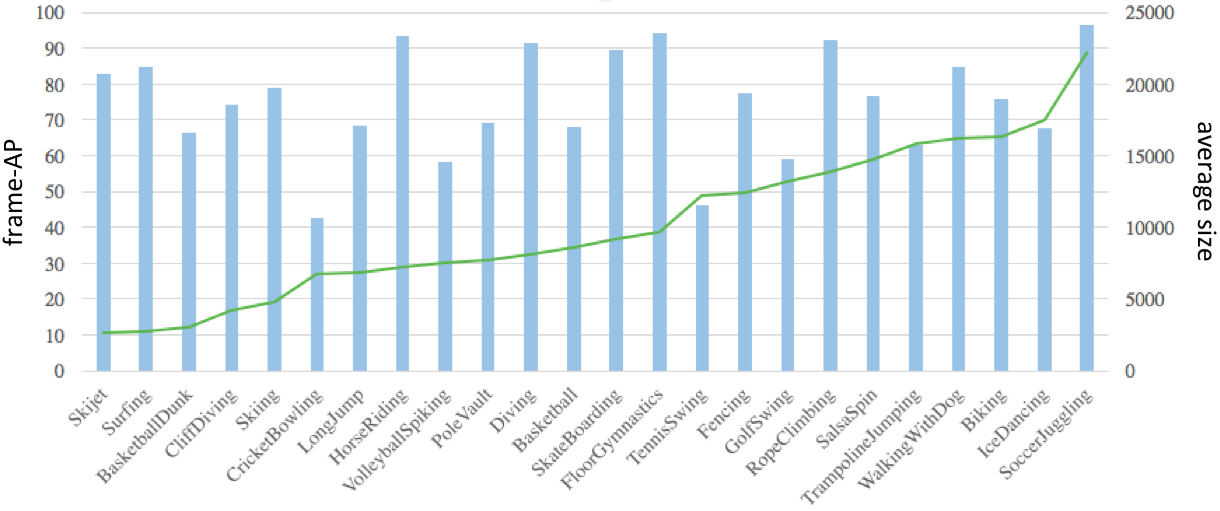}
\end{center}
\vspace{-.2in}
\caption{Analysis of the detection accuracy (blue) and the average bounding box size (green) of each action class.}
\label{fig:accuracy_size}
\end{figure}

\begin{figure}[t]
\begin{center}
\includegraphics[width=\columnwidth]{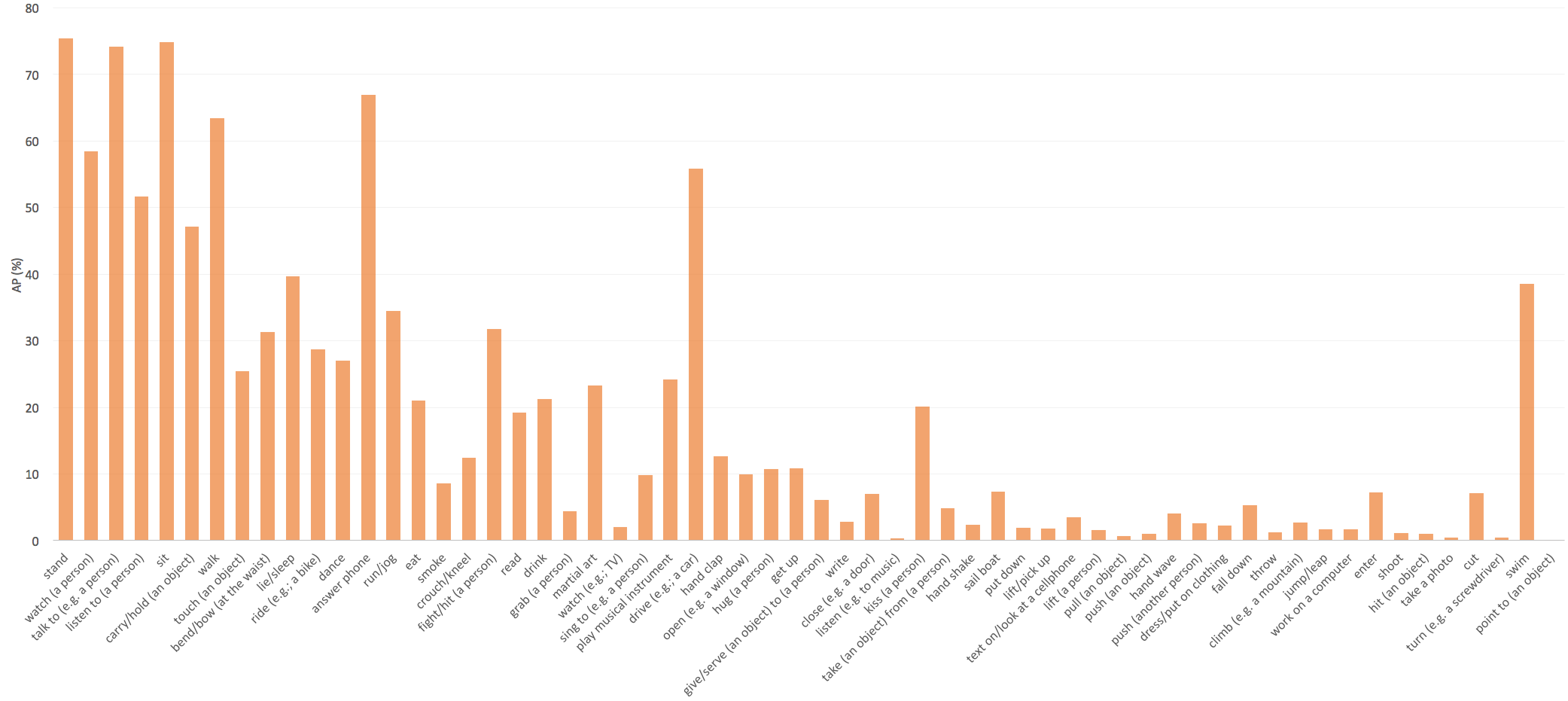}
\end{center}
\vspace{-.2in}
\caption{Comparison of the per-class breakdown frame-AP at IoU threshold $0.5$ on AVA.}
\label{fig:per-class}
\end{figure}

In order to tackle the spatial displacement problem, we introduce two methods to adaptively perform the temporal extension, i.e., extrapolation and anticipation as defined in Eqs.(4-5) of the paper. Figure~\ref{fig:extrapolate} illustrates the extrapolation: for each of the current proposals, following its first and last tubelets (one tubelet with 6 bounding boxes), the extrapolation linearly estimates the directions and scales of the extended tubelets. As for the impact of different action scales, we qualitatively show the examples in Figure 8 of the paper, and we report the frame-APs and average sizes of different action classes of UCF101 in Figure~\ref{fig:accuracy_size}. Thanks to the progressive learning, STEP is found to be robust to handle the actions with small scales, though it starts with coarse-scale proposals. Figure~\ref{fig:per-class} demonstrates the per-class breakdown frame-AP on AVA.  

\end{document}